# Transform, Contrast and Tell: Coherent Entity-Aware Multi-Image Captioning

Jingqiang Chen
School of Computer Science, Nanjing University of Posts and Telecommunications, Nanjing, China
cjq@njupt.edu.cn

**Abstract**

Coherent entity-aware multi-image captioning aims to generate coherent captions for neighboring images in a news document. There are coherence relationships among neighboring images because they often describe same entities or events. These relationships are important for entity-aware multi-image captioning, but are neglected in entity-aware single-image captioning. Most existing work focuses on single-image captioning, while multi-image captioning has not been explored before. Hence, this paper proposes a coherent entity-aware multi-image captioning model by making use of coherence relationships. The model consists of a Transformer-based caption generation model and two types of contrastive learning-based coherence mechanisms. The generation model generates the caption by paying attention to the image and the accompanying text. The caption-caption coherence mechanism aims to render entities in the caption of the image be also in captions of neighboring images. The caption-image-text coherence mechanism aims to render entities in the caption of the image be also in the accompanying text. To evaluate coherence between captions, two coherence evaluation metrics are proposed. The new dataset DM800K is constructed that has more images per document than two existing datasets GoodNews and NYT800K, and is more suitable for multi-image captioning. Experiments on three datasets show the proposed captioning model outperforms 7 baselines according to BLUE, Rouge, METEOR, and entity precision and recall scores. Experiments also show that the generated captions are more coherent than that of baselines according to caption entity scores, caption Rouge scores, the two proposed coherence evaluation metrics, and human evaluations.

## 1 Introduction

There are a large number of images in the Internet, many of which do not have proper captions. A great body of research on generic image captioning have been carried out to generate common captions describing everyday objects and their relationships [1,2,3]. Recently developed entity-aware image captioning aims to generate specific informative captions that describe named entities and events in the images using information in accompanying news documents [4,5,6,7,8]. Entity-aware image captioning or news image captioning is different from traditional generic image captioning mainly in the following two aspects. Firstly, news images are placed in new text which should be considered for caption generation. Secondly, news image captions contain more fine-grained information such as entities than generic captions do. However, current entity-aware image captioning focuses on single-image captioning, which treats every image independently and neglects coherence relationships between neighboring images. These coherence relationships are important for entity-aware multi-image captioning, which is different from entity-aware single-image captioning and has not been explored before.

There are coherence relationships between captions of neighboring images in a document, because neighboring images often describe same or related events and entities from different aspects. Fig. 1 shows an example taken from Daily Mail. This example is about Nasa's launching of Juno probe to the outer solar system. There are three coherent images in the news document. The first image is about the launching scene of Juno probe, which is launched by Nasa to gather information about Jupiter. The following two images are about the scenes after the probe is launched. Concretely, the second image contains a windmill-shaped object describing the shape of Juno probe with three solar panels, and the third image describes the size of the solar panels in more details. The captions of the three neighboring images share common entities and nouns, and are coherent with each other. This type of coherence is called caption-caption coherence in this paper, rendering the entities mentioned in the generated caption of an image be also in captions of its neighboring images. For example, the entities Juno spacecraft, the solar system and solar panels occur in all the captions of the images, and thus connects the images. Caption-caption coherence has not been explored by current entity-aware image captioning methods.

Most existing work [4,5,6,7,8] on entity-aware image captioning focuses on single-image captioning, which aims to create a caption of a news image that is coherent with its accompanying text and the news image. This type of coherence is called caption-image-text coherence in this paper, rendering the entities mentioned in the accompanying text of the image be also in the caption. For example, the entities and the nouns spacecraft, Juno, Jupiter in the caption of the first image also occur in the surrounding text. As with caption-image-text coherence, some studies [9,10,11] make use of image-text relations for generic single-image captioning by capturing the structural, logical, and purposeful relationships between the visual and textual modality. Other work [12] generates a paragraph as the image caption and measures coherence as similarity between two neighboring sentences in the paragraph. These studies on coherence of image captioning are all for single-image captioning.

For the task of entity-aware multi-image captioning, both caption-caption coherence and caption-image-text coherence should be considered to generate coherent captions for multiple images in a document. Entity-aware multi-image captioning is a novel task that has not been explored before. Caption-caption coherence is important for the task.

To fill the gap, this paper proposes the **coh**erent entity-aware multi-image **cap**tioning model (CohCaps for short), utilizing caption-image-text coherence and caption-caption coherence to generate coherent captions of neighboring image in a document. Contrastive learning is employed to model and learn the two types of coherence by maximizing a given anchor point's similarity to a "positive" sample and minimizing a given anchor point's similarity to a "negative" sample. Contrastive learning has been successfully used in many applications including image captioning recently [3,13]. CohCaps consists of a Transformer-based generation model and two contrastive coherence mechanisms. (i) The generation model generates captions from news images and accompanying texts, because information for news image captioning is not only contained in the image but also contained in the associated news articles. In the encoding stage, texts are encoded with Roberta [14] to extract text representations. Images are encoded to extract image representations, object representations, and face representations. In the decoding stage, four types of representations are fused for caption generation. Coherence mechanisms work in the decoding stage of the generation model. (ii) The contrastive caption-image-text coherence mechanism is proposed to model caption-image-text coherence by contrasting the ground-of-the-truth caption with the fake caption of the image, rendering the generated caption be coherent with its surrounding text. The fake caption is created by replacing entities in the true caption with randomly selected entities. (iii) The contrastive caption-caption coherence mechanism is proposed to model caption-caption coherence, rendering the generated caption be coherent with captions of neighboring images. The positive sample is created by pairing the ground-of-the-true caption of the image with that of neighboring images in the same document. To create negative samples, two approaches are proposed. The first approach is to pair the ground-of-the-truth caption of the image with the fake caption of the neighboring image. The second approach is to pair ground-of-the-truth captions of images in different documents as the negative samples. (iv) To evaluate coherence between captions of neighboring images, two coherence metrics are proposed based on caption-caption coherence mechanisms.

**Nasa to launch Juno probe to Jupiter just two weeks after end of shuttle programme**

Just two weeks after the end of its historic shuttle programme, Nasa will today launch a solar-powered spacecraft on a mission to Jupiter. The robotic explorer Juno is set to become the most distant probe ever powered by the sun. The windmill-shaped craft is equipped with three tractor-trailer-size solar panels for its two billion-mile journey into the outer solar system. It will blast-off from Cape Canaveral in Florida at 11.34am local time (4.34pm BST) aboard an unmanned Atlas V rocket.

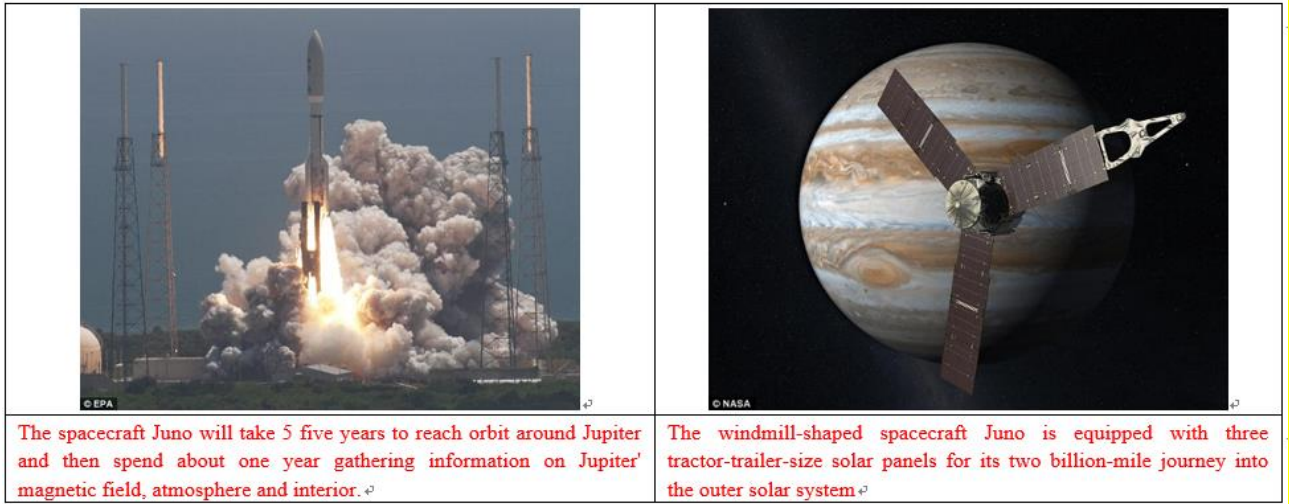

| The spacecraft Juno will take 5 five years to reach orbit around Jupiter and then spend about one year gathering information on Jupiter' magnetic field, atmosphere and interior. | The windmill-shaped spacecraft Juno is equipped with three tractor-trailer-size solar panels for its two billion-mile journey into the outer solar system |

After the landing of shuttle Atlantis prompted waves of nostalgia around the world, the demise of that Nasa programme is giving extra oomph to the $1.1billion voyage to the largest and probably oldest planet in the solar system.

It is the first of three high-profile astronomy missions coming up for Nasa in the next four months. Jupiter - a planet several Nasa spacecraft have studied before - is so vast it could hold everything else in the solar system, minus the sun.

Scientists hope to learn more about planetary origins through Juno's exploration of the giant gas-filled planet, a body far different from rocky Earth and Mars. Jim Gree, Nasa's director of planetary science, said: It is a new era.

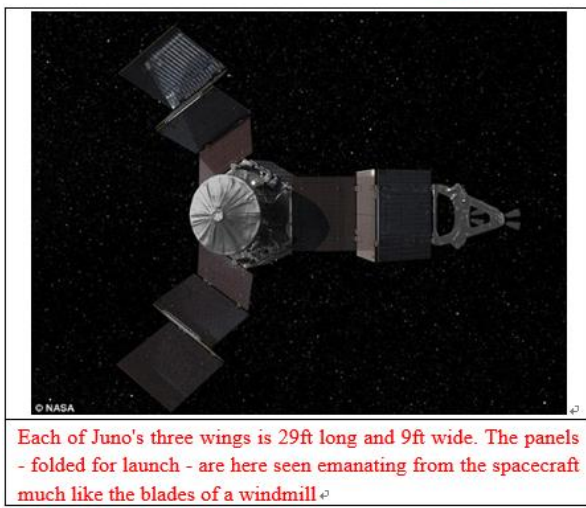

Each of Juno's three wings is 29ft long and 9ft wide. The panels - folded for launch - are here seen emanating from the spacecraft much like the blades of a windmill

'Humans plan to go beyond low-Earth orbit. When we do that, it's not like Star Trek. It's not "go where no man has gone before".' he said.

It will take Juno five years to reach its target, five times farther from the sun than Earth. No spacecraft has ever ventured so far, powered by solar wings. Europe's solar-powered, comet-chasing Rosetta probe made it as far as the asteroid belt between the orbits of Mars and Jupiter. Each of Juno's three wings is 29ft long and 9ft wide, necessary given that Jupiter receives 25 times less sunlight than Earth. The panels - folded for launch - emanate from the spacecraft much like the blades of a windmill. At Jupiter, nearly 500million miles from the sun, Juno's panels will provide 400 watts of power. In orbit around Earth, these panels would generate 35 times as much power.

Fig. 1　An example of multi-image captioning taken from Daily Mail

Since existing datasets for entity-aware image captioning are mainly for single image captioning and the average number of images in documents are small, this paper constructs the dataset named DM800K for multi-image captioning. DM800K is constructed by collecting news documents and news images from the DailyMail website. DM800K contains 4.5 images per document, much larger than that of existing datasets NYTimes800K [4] and GoodNews [5].

The main contributions of this paper can be summarized as follows:

- This paper explores the task of entity-aware multi-image captioning, and proposes the captioning model named CohCaps with a caption-image-text coherence mechanism and two caption-caption coherence mechanisms to generate coherence captions for neighboring images in an article.
- This paper constructs the new dataset DM800K by collecting images, captions and documents from the Daily Mail website. DM800K contains more images per document and is more suitable for multi-image captioning.
- Experiments on three datasets show that the proposed method outperforms 7 SOTA methods according to single-image captioning evaluations, and show that the captions generated by the proposed method are more coherent than that of baselines according to automatic evaluations and human evaluations.

## 2 Related work

**Image captioning** is a multi-modal task that is relevant to natural language understanding and computer vision, and has gained much attention from the two research areas. Modern image captioning models are based on deep learning techniques [15,16]. Most recent progress made on generic image captioning directly generate captions from images based on the encoder-decoder model, using the convolution neural network to encode the image for image representations, and using the recurrent neural network or Transformer [17] to generate the caption. Vinyals et al. [15] proposes to attend to different image patches in the decoding steps, and Lu et al. [16] follows by not attending to image regions sometimes. You et al. [18] imposes local representations at the object level and a bottom-up mechanism [19] to combine salient image regions. Contrastive learning is also applied to image captioning to generate distinctive captions [20].

**Entity-aware image captioning** or news image captioning is different from traditional generic image captioning. Existing entity-aware image captioning research focuses on single-image captioning. Chen and Zhuge [8] proposes a multi-modal pointer-generator network for news image captioning with the multimodal pointer mechanism which uses both textual attention and visual attention to compute pointer distributions, and with the multimodal coverage mechanism to reduce repetitions of attentions or repetitions of pointer distributions. Biten et al. [5] adopts a template-based encoder-decoder method which extracts named entities by paying attention to sentences of the accompanying text, and proposes the dataset named GoodNews for the task. Tran, Mathews, and Xie [4] follow by proposing the dataset NYTimes800K, and by proposing the Transformer-based news image captioning method in an end-to-end manner. Zhao and Wu [7] follow to construct multi-modal knowledge graph which is added into the transformer model to improve entity-aware image captioning. Liu et al. [6] propose another type of news image captioning method which effectively combine visual and textual features to generate caption by utilizing much fewer parameters while achieving slightly better prediction results. The task of entity-aware multi-image captioning has not been explored before. A related study is group-based image captioning which aims to generate captions for a group of related images to reflect diversity of the images [20]. Entity-aware multi-image captioning is different from group-based image captioning mainly in that the images for the former task are in a same document, so the task has to consider accompanying text and coherence information of the images, while the latter task does not.

**Coherence mechanisms** have been studied and applied to many NLP tasks and obtain good performance, such as single-image captioning [9,10,12,21], text summarization [22,23], gesture interpretation [25,26], text-to-image retrieval [11], and machine comprehension [26]. Recent studies on coherence for single-image captioning are based on the framework of discourse coherence theory [27] which uses a constrained inventory of coherence relations. Alikhani et al. [9] proposes the new image-text relations for single-image captioning that capturing the structural, logical, and purposeful relationships between the visual modality and the textual modality. They categorize image-text coherence relations as Visible relations, Restatement relations, Story relations, and Occasion relations, which are used to control single-image captioning models. Inan et al. [10] follow to propose a coherence-aware metric for single-image captioning by constructing a dataset of image-description pairs that are annotated with coherence relations. And Alikhani et al. [11] follow to apply this type of coherence relations to the text-to-image retrieval task. Another type of coherence is based on the similarity measure. He and Li [12] propose to generate a paragraph as the image caption by modeling coherence and diversity of the sentences in the generated paragraph which is used as the reward in the reinforcement learning. They measure coherence as the similarity between two neighboring sentences in the paragraph. Similarly, Wu and Hu [28] use contrastive learning to trains the coherence model for text summarization using neighboring sentences as positives samples and non-neighboring sentences as negative samples. And then Sharma et al. [22] follow to propose to incorporate coherence into entity-driven text summarization by using such type of computed coherence as the reward of self-critic reinforcement learning to generate coherent summaries. The coherence mechanisms for entity-aware multi-image captioning proposed in this paper are different from that of previous work in two aspects. Firstly, the proposed coherence mechanisms are entity-aware coherence mechanisms, which compute coherence based on the entities on the captions and the accompanying texts. Secondly, the proposed coherence mechanisms not only consider coherence between the caption and the accompanying text, but also consider coherence between captions of neighboring images.

## 3 The dataset DM800K

### 3.1 Construction of DM800K

Since the task of multi-image captioning aims to generate coherent captions for multiple images, the input document should contain more than one image. The existing datasets GoodNews and NYTimes800K are initially created for entity-aware single-image captioning by collecting text and images from the New York Times website, and contain relatively less images per document. Therefore, we create the new dataset DM800K that contains more images per document. DM800K can be more suitable for the task of multi-image captioning.

**Table 1** General Statistics of GoodNews, NYTimes800K, and DM800K

|  | GoodNews | NYTimes800K | DM800K |
| --- | --- | --- | --- |
| **Average number of images per document** | **1.80** | **1.78** | **4.45** |
| **Number of images** | 462, 642 | 792, 971 | 813, 476 |
| **Number of documemts** | 257, 033 | 444, 914 | 182, 756 |
| **Average Document Length** | 451 | 974 | 578 |
| **Average Caption Length** | 18 | 18 | 22 |
| **% of caption words that are** |  |  |  |
| - nouns | 16% | 16% | 18% |
| - verbs | 9% | 4% | 15% |
| - pronouns | 1% | 1% | 2% |
| - proper names | 23% | 22% | 14% |
| - adjectives | 4% | 4% | 5% |
| - named entities | 18% | 26% | 18% |
| **% of captions with named entities** | 97% | 96% | 70% |

For sake of efficiency, DM800K is created based on the dataset DailyMail [29] which is collected from the Daily Mail website and has been widely used for text summarization. The DailyMail dataset provides the html-formatted documents, which can be parsed to extract image captions, image URLs, image positions, and news text. Images are further downloaded via the extracted image URLs. Because the positions of the images is given, the surrounding 512 words of each image can be extracted from the news text, and the order of the images in a document can also be determined via image positions. Since most of the documents in the original DailyMail dataset contain 1 to 10 images, these documents are selected in DM800K for the task of multi-image captioning. The splits of train, dev, and test in DM800K are the same as DailyMail. MongoDB is adopted to store DM800K.

Table 1 shows statistics of the three datasets. The first line of the table shows that the average numbers of images per document of the three datasets are 1.80, 1.78, and 4.45 respectively. Documents in DM800K have nearly twice more images than documents in GoodNews and NYTimes800K. Other statistics of DM800K is also shown in the table such as percent of different types of words, named entities, average document length and caption length, which is not very different from that of GoodNews and NYTimes800K.

**3.2 Discussion of coherence of neighborhood images**

We argue that two neighborhood images in a news document are coherent with each other, because they have similar text context describing same or related events or entities. And accordingly, the corresponding captions are also coherent with each other. To support the argument, we compute rouge scores of text contexts, rouge scores and entity coverage scores of captions for different hops of neighborhood images. The scores for 1 to 5-hops neighborhood images are listed in Table 2.

In Table 2, 1-hops neighborhood images are two directly neighboring images, 2-hops neighborhood images are two neighboring images with one image hop, and so on. To measure coherence of neighboring images, the following three scores can be computed.

**Text Rouge Score** measures text overlaps of accompanying text of two neighborhood images, and can be computed as in equation (1). In the equation, $Rouge(ref, sys)$ is the function for computing Rouge-L scores [30] where $ref$ is the reference text and $sys$ is the text being evaluated, and $txt_1$ and $txt_2$ denote the accompanying texts of two neighboring images respectively.

$$TextRougeScore(txt_1, txt_2) = \frac{Rouge(txt_1, txt_2) + Rouge(txt_2, txt_1)}{2} \quad (1)$$

**Caption Rouge Score** measures text overlaps of captions of two neighboring images, and can be computed as in equation (2). In the equation, $cap_1$ and $cap_2$ denote the captions of two neighboring images respectively

$$CaptionRougeScore(cap_1, cap_2) = \frac{Rouge(cap_1, cap_2) + Rouge(cap_2, cap_1)}{2} \quad (2)$$

**Caption Entity Score** measures common entities contained in captions of two neighboring images, and is computed as in equation (3) to (5). In the equations, es$_1$ and es$_2$ denote the entity sets contained in the captions of the two neighboring images respectively. The calculation method of Caption Entity Score is similar to that of F1-Score, using es$_1$ and es$_2$ as the reference and the hypothesis alternatively.

$$P(es_1, es_2) = \frac{|es_1 \cap es_2|}{|es_1|} \quad (3)$$

$$R(es_1, es_2) = \frac{|es_1 \cap es_2|}{|es_2|} \quad (4)$$

$$CaptionEntityScore(es_1, es_2) = \frac{2 \times P(es_1, es_2) \times R(es_1, es_2)}{P(es_1, es_2) + R(es_1, es_2)} \quad (5)$$

**Table 2** Rouge scores and entity scores of N-hops neighborhood images in GoodNews, DM800K and NYTimes800K

| Dataset | N-hops Neighborhood Images | Text Rouge Score | Caption Rouge Score | Caption Entity Score |
|---|---|---|---|---|
| DM800K | 1-hops | 91.70 | 27.37 | 0.2770 |
|  | 2-hops | 84.12 | 15.47 | 0.1755 |
|  | 3-hops | 77.71 | 13.61 | 0.1570 |
|  | 4-hops | 72.11 | 11.95 | 0.1391 |
|  | 5-hops | 67.39 | 11.57 | 0.1332 |
| NYTimes800K | 1-hops | 55.62 | 13.23 | 0.1342 |
|  | 2-hops | 37.47 | 12.46 | 0.1116 |
|  | 3-hops | 32.89 | 12.04 | 0.0989 |
|  | 4-hops | 32.83 | 12.27 | 0.0953 |
|  | 5-hops | 34.39 | 12.54 | 0.0901 |
| GoodNews | 1-hops | - | 13.36 | 0.1341 |
|  | 2-hops | - | 12.55 | 0.1113 |
|  | 3-hops | - | 11.95 | 0.0982 |
|  | 4-hops | - | 11.87 | 0.0916 |
|  | 5-hops | - | 11.94 | 0.0877 |

As shown in Table 2, Text Rouge Scores, Caption Rouge Scores and Caption Entity Scores decrease when hops of images increase. More distant two images are, less common information they share. Text Rouge Scores for the GoodNews dataset are not reported in the table, because image positions are not provided in the dataset. Therefore, the first 512 words are used as text context instead for GoodNews. The scores of DM800K are much higher than that of NYTimes800K and GoodNews, indicating that neighboring images in DM800K share more common information. For all the three datasets, scores for 1-hop neighboring images are highest, indicating 1-hop neighboring images share the most common information. Therefore, we use 1-hop neighboring images i.e. two directly neighboring images to train our coherence mechanisms in the following.

## 4 The proposed model

The goal of the proposed **coh**erent entity-aware multi-image **cap**tioning model (CohCaps for short) is to generate coherent captions for neighboring images in a document by utilizing caption-image-text coherence and caption-caption coherence. As shown in Fig. 2, the proposed framework of the multi-image captioning model consists of the Transformer-based caption generation model and the Contrastive Learning-based coherence mechanisms. The caption generation model uses four encoders to extract representations of text and images, and uses the Transformer decoder and multi-head attentional mechanisms to generate captions. The generation model is equipped with two types of contrastive coherence mechanisms. The caption-image-text coherence mechanism is to render entities in the generated caption be also in the accompanying text. The caption-caption coherence mechanisms are to render entities in the generated caption be also in captions of neighboring images in the same document.

$$B = \{(Img_i, Txt_i, TrueCap_i, FakeCap_i, DocIndex_i, ImgIndex_i) \mid i \in [1, BatchSize]\} \quad (6)$$

A batch of input is a set of images with accompanying texts denoted as in equation (6), where $Img_i$ is the image, $Txt_i$ is the accompanying text, $TrueCap_i$ is the ground-of-the-truth caption, $FakeCap_i$ is the fake caption created by replacing entities in $TrueCap_i$ with randomly selected same-typed entities, $DocIndex_i$ is the index of the document that the image is from, and $ImgIndex_i$ is the index of the image in the document. For example, the fake caption of the first image in Fig. 2 is created by replacing the entities Juno and Jupiter in the ground-of-the-truth caption with the same-typed entities Anna and Mars respectively. The fake caption is not coherent with the image, the accompanying text and the neighboring images. The caption-image-text coherence mechanism contrast $TrueCap_i$ with $FakeCap_i$ of the image. As shown in Fig. 2, there are two types of caption-caption coherence mechanisms. The first caption-caption coherence mechanism contrasts the true caption of an image with the fake captions of the neighboring images from the same document, and the second caption-caption coherence mechanisms contrasts the true caption of an image with true captions of images from different documents. $DocIndex_i$ is used to determine whether the images are from a same document, and $ImgIndex_i$ is used to determine whether the images are neighboring images. The special token <S> is prepended to the beginning of the caption, and the special token </S> is appended to the end of the caption. Since the token </S> is in the end of the caption, the decoding state corresponding to the token can be used for contrasting.

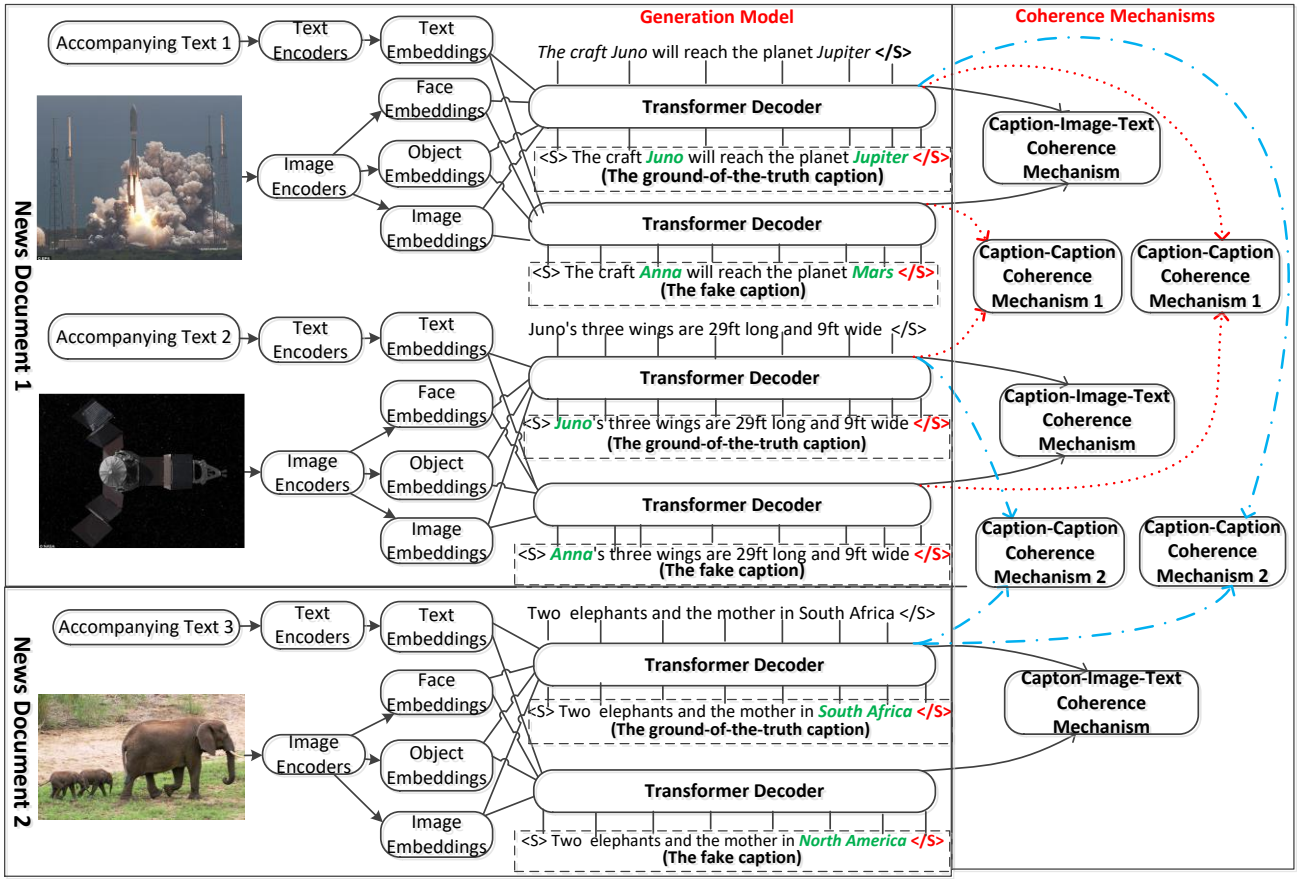

**Fig. 2** The framework of the proposed model, consisting of a caption generation model, a caption-image-text coherence mechanism, and two caption-caption coherence mechanisms. Each image is paired with a ground-of-the-truth caption and a fake caption. The fake caption is created by replacing entities in the ground-of-the-truth caption with randomly selected same-typed entities.

For a batch of input, there are several groups of images from different documents. For a document, at most *W* neighboring images are selected in a batch. Suppose the batch size is set as 15, and *W* is set as 3. There can be 5 groups of images in the batch. Images in different group are from different documents. Take Fig. 2 for example. The first image and the second image in Fig. 2 are from a same document, and the third image in Fig. 2 is from a different document. The details of the generation model and the coherence mechanisms are introduced in the following.

**4.1 The caption generation model**

The Transformer-based single-image captioning model is borrowed from [4]. It consists of four encoders and a decoder. The encoders encode the accompanying text to obtain text representations, and encode the image to obtain image encodings, object embeddings, and face embeddings. The decoder generates the caption by paying attention to the four types of encodings using the multi-head attentional mechanism and the self-attentional mechanism.

**Encoders** Four encoders are adopted to encode text and images. (i) The first encoder is the text encoder which uses RoBERTa to encode the accompanying text of the image. Words in the accompanying text are first encoded as byte pairs, and are then fed to RoBERTa with 24 encoding layers. Encodings of a byte token in each layer are weighted summed as the final encoding of the byte token for the decoder to attend to. These encodings are denoted as $X^{Txt} \in R^{L^{Txt} \times d^{Txt}}$, where $L^{Txt}$ is the length of text and $d^{Txt}$ is the dimension of the embeddings set as 2048. (ii) The second encoder is the image encoder which uses ResNet-152 [31] pre-trained on ImageNet to encode images. The output of the final block before the pooling layer is used as the image representation. These encodings are denoted as $X^{Img} \in R^{49 \times d^{Img}}$, where 49 is the number of image blocks and $d^{Img}$ is the dimension of the embeddings set as 2048. (iii) The third encoder is the face encoder which uses MTCNN [32] to detect face bounding boxes in the image, and uses the FaceNet [33] pre-trained on the VGGFace2 dataset [34] to encode the face bounding boxes. These encodings are denoted as $X^{Face} \in R^{4 \times d^{Face}}$, where 4 is the 4 selected top-ranked faces and $d^{Face}$ is the dimension of the embedding set as 512. (iv) The fourth encoder is the object encoder which uses YOLOv3 [35] to detect object bounding boxes in the image and then uses through a ResNet-152 to encode the objects. These encodings are denoted as $X^{Obj} \in R^{L^{Obj} \times d^{Obj}}$, where $L^{Obj}$ is the number of objects selected in the image and $d^{Obj}$ is the dimension of the embeddings set as 2048.

**Decoder** The decoder of the captioning model consists of L identical Transformer layers and generates caption tokens sequentially. The context to the decoder is denoted as $X = [X^{Txt}; X^{Img}; X^{Face}; X^{Obj}]$, where the operator [;] denotes the concatenation operation. At the *t*-th time step, the decoder predicts the probability of the current token $c_t \in Vocab$ using the context and the embeddings of the previously generated caption words, where $c_t$ denotes the *t*-th caption word and *Vocab* denotes the vocabulary. The details are as follows.

Suppose $s_{l,t}$ is the output of the $\ell$-th transformer decoder layer at the time step *t*, which is obtain by using the masked self-attention mechanism [17] to attend to the past tokens as in equation (7) and (8), and using the multi-head attention mechanism [17] to attend to the multi-modal encodings *X* as in equation (9).

$$s_{0,*} = [<S>, c_1, c_2, \cdots, </S>] \quad (7)$$

$$\tilde{s}_{l,*} = MaskedSelfAtt(s_{l-1,*}) \quad (8)$$

$$s_{l,*} = MHAtt(\tilde{s}_{l,*} \mid X^{Txt}, X^{Img}, X^{Face}, X^{Obj}) \quad (9)$$

In equation (7), $s_{0,*}$ is the caption word sequence of the initial layer, which is set as the ground-of-the-truth caption words for training. In equation (8), *MaskedSelfAtt* denotes the masked self-attentional mechanism, and is used to compute $\tilde{s}_{l,*}$ by paying attention to the previous generated tokens. For sake of computational efficiency, the dynamic convolutions [36] can be adopted to replace the self-attentional mechanism as in [4]. In equation (9), *MHAtt* denotes the multi-head attentional mechanism, and is used to compute $s_{l,*}$ by paying attention to $X^{Txt}$, $X^{Img}$, $X^{Face}$, and $X^{Obj}$ using $\tilde{s}_{l,*}$ as queries.

$$p(y_t) \sim soft\max(s_{L,t}) \quad (10)$$

$$loss^{Gen} = -\sum_i \sum_t \log p(y_t^{(i)}) \quad (11)$$

Finally, equation (10) is used to estimate $p(y_t)$, the probability of generating the *t*-th token in the vocabulary via *softmax*. And equation (11) is used to calculate the generative loss by summing up negative log likelihood of the generating probability of the target word for all the training samples. In equation (11), $y_t^{(i)}$ denotes the *t*-th generated caption word for the *i*-th training sample.

Let $s_{L,</S>}$ denotes the state for the special token </S> in the *L*-th layer. The state $s_{L,</S>}$ can accumulate information of the caption, the image and the accompanying text through the self-attention mechanism and the multi-head attention mechanism. This state can be used to contrast different captions in the following defined coherence mechanisms.

### 4.2 The caption-image-text coherence mechanism

The caption-image-text coherence mechanism (CapITCoh for short in the following) is proposed to make the generated caption be coherent with the accompanying text and the image. Concretely, if the caption is coherent with the accompanying text, the entities in the generated caption will be relevant with entities in the accompanying text. Otherwise, if the caption is not coherent with the text, the entities in the generated caption will be not relevant with the accompanying text. Hence, contrastive learning is employed to model and learn the caption-image-text coherence. The first problem is to create positive samples and negative samples for training.

**Positive samples for CapITCoh** (True captions) The ground-of-the-truth caption is used as the positive samples. The special sequence starting token <S> are prepended to the caption, and the sequence ending token </S> is appended to the caption. This positive sample is also called the true caption in the following. Then the positive sample is denoted as $TrueCap_i = [<S>, c_{i,1}^{True}, c_{i,2}^{True}, \cdots, </S>]$.

**Negative samples for CapITCoh** (Fake captions) The negative samples are created by replacing the entities in the ground-of-the-truth captions with randomly selected same-typed entities. Here we can simply use the entity labels recognized by the SpaCy NER model as the entity types. For example, the entities the spacecraft Juno and the planet Jupiter in the ground-of-the-truth caption of the first image in Fig. 1 are labeled as PERSON and LOC respectively by SpaCy, and they can be replaced with Anna and Mars which are also labeled as PERSON and LOC respectively. This negative sample is also called the fake caption in the following. The fake caption is denoted as $FakeCap_i = [<S>, c_{i,1}^{Fake}, c_{i,2}^{Fake}, \cdots, </S>]$.

**Contrastive loss for CapITCoh** Contrastive learning is employed to model caption-image-text coherence by contrasting positive samples with negative samples. The difference between true captions and fake captions is the entities in the captions. True captions are coherent with the accompanying text and the image, while fake captions are not coherent with the text and the image. As mentioned in the previous subsection, the decoding state $s_{L,</S>}$ for the special token </S> in the *L*-th layer can fuse information of the caption, the image and the accompanying text. There are a positive sample and a fake caption for every image. Hence, the contrastive loss for CapITCoh can be defined based on the decoding state $s_{L,</S>}$ of the true caption and the fake caption as in equation (12).

$$loss^{CapITCoh} = \sum_i -\log \sigma(MLP^{CapITCoh}(s_{L,</S>}^{True^{(i)}})) - \log(1 - \sigma(MLP^{CapITCoh}(s_{L,</S>}^{Fake^{(i)}}))) \quad (12)$$

In equation (12), $s_{L,</S>}^{True^{(i)}}$ denotes the decoding state $s_{L,</S>}$ for the true caption of the *i*-th image, and $s_{L,</S>}^{Fake^{(i)}}$ denotes the decoding state $s_{L,</S>}$ for the fake caption of the *i*-th image. A two-layer MLP is used to compute logits for $s_{L,</S>}^{True^{(i)}}$ and $s_{L,</S>}^{Fake^{(i)}}$, and the sigmoid function is used as the activation function. And then Cross Entropy is employed to compute the contrastive loss $loss^{CapITCoh}$.

### 4.3 The first caption-caption coherence mechanism CapCapCoh1

The caption-caption coherence mechanism (CapCapCoh for short) is proposed to make the generated caption of the image coherent with captions of neighboring images in the same document. Concretely, if the caption of an image is coherent with the caption of another image in the same document, the entities in the former caption will also occur in the latter caption. This is because these two captions in the same document describe related events or topics, especially when these images are neighboring images. Otherwise, if the caption is not coherent with captions of other images, the entities in the former caption will be not relevant with the entities in captions of other images. Contrastive learning is employed to learn caption-caption coherence. Positive samples for the caption-caption coherence mechanism are constructed by pairing true captions of an image. To create negative samples, there are two approaches, corresponding to two different caption-caption coherence mechanisms. This subsection introduces the first one named CapCapCoh1, and the next subsection introduces the second one named CapCapCoh2.

**Positive samples for CapCapCoh** Each positive sample contains true captions of two neighboring images from a same document. Such two captions are coherent with each other because their corresponding images are related to each other and are about a same event or topic. Intuitively, more adjacent two images in a document are, more coherent the corresponding captions are. Take the first two images in Fig. 2 for example. The two images are neighboring images from a document. The captions of the first image and the second contain the same entity spacecraft, with the first image describing the launching of the craft and the second image describing the equipped panels of the craft. To constraints adjacency of images, at most $W$ neighboring images (empirically set as 3 in the following) are selected from a document in a batch as mentioned. Then the positive sample for CapCapCoh is created by pairing true captions of neighboring images within the window $W$ in a batch as in equation (13).

$$CapCapPos = \{(TrueCap_i, TrueCap_j) \mid DocIndex_i = DocIndex_j \land |ImgIndex_i - ImgIndex_j| = 1\} \quad (13)$$

In the equation (13), TrueCap$_i$ and TrueCap$_j$ are the true captions of the $i$-th image and the $j$-th image respectively, DocIndex$_i$ and DocIndex$_j$ are the document indices of the $i$-th image and the $j$-th image respectively, and ImgIndex$_i$ and ImgIndex$_j$ are the image indices of the $i$-th image and the $j$-th image respectively. The condition $|ImgIndex_i - ImgIndex_j| = 1$ restricts that the two images are 1-hop neighboring images. For every two images in the window, a positive sample for CapCapCoh is created.

**Negative samples for CapCapCoh1** In contrast to positive samples, entities in negative samples are not coherent. CapCapCoh1 creates negative sample CapCapNeg1 by pairing the true caption of one image with the fake caption of another image in the same document as in equation (14).

$$CapCapNeg1 = \{(TrueCap_i, FakeCap_j) \mid DocIndex_i = DocIndex_j \land |ImgIndex_i - ImgIndex_j| = 1\} \quad (14)$$

Equation (14) is similar to equation (13), with the difference that TrueCap$_j$ in equation (13) is replaced with FakeCap$_j$ in equation (14). For every two 1-hop neighboring images $i$ and $j$, two negative samples can be created by paing TrueCap$_i$ with FakeCap$_j$ and by pairing TrueCap$_j$ with FakeCap$_i$.

Captions in the positive sample are coherent with each other, while captions in the negative sample are not coherence with each other. Since the positive sample and the negative samples created from the two neighboring images $i$ and $j$ have the same contexts of text and images, these samples can be contrasted to learn coherence between the captions of neighboring images. Accordingly, equation (15) is defined to compute the contrastive loss **loss$^{CapCapCoh1}$**.

$$loss^{CapCapCoh1} = - \sum_{(TrueCap_i, TrueCap_j) \in CapCapPos} \log \sigma(MLP^{CapCapCoh1}([s_{L,</S>}^{True^{(i)}}, s_{L,</S>}^{True^{(j)}}]))$$
$$- \sum_{(TrueCap_i, FakeCap_j) \in CapCapNeg1} (1 - \log \sigma(MLP^{CapCapCoh1}([s_{L,</S>}^{True^{(i)}}, s_{L,</S>}^{Fake^{(j)}}]))) \quad (15)$$

In equation (15), $s_{L,</S>}^{True^{(i)}}$, $s_{L,</S>}^{True^{(j)}}$ and $s_{L,</S>}^{Fake^{(j)}}$ denote the decoding states $s_{L,</S>}$ for the captions TrueCap$_i$, TrueCap$_j$, and FakeCap$_j$ respectively. The notation [] in the equation denotes the concatenation operation. The states $s_{L,</S>}^{True^{(i)}}$ and $s_{L,</S>}^{True^{(j)}}$ for the positive sample are concatenated and a two-layer MLP is applied to compute the logits for the positive samples. The states $s_{L,</S>}^{True^{(i)}}$, $s_{L,</S>}^{Fake^{(j)}}$ for the negative sample of CapCapCoh1 are concatenated to compute the logits. And then the Cross Entropy is employed is used to compute the loss of CapCapCoh1.

**4.4 The second caption-caption coherence mechanism CapCapCoh2**

The positive samples for CapCapCoh2 is the same as that of CapCapCoh1. The negative samples of CapCapCoh2 are created by pairing the true captions of two images from different documents. Captions in the positive sample has the same text and image context, and contains related entities. While captions in the negative sample CapCapNeg2 has the different text and image context, or even the words of the captions are totally different. These positive and negative samples are contrasted to enhance coherence of captions under the same context.

$$CapCapNeg2 = \{(TrueCap_i, TrueCap_j) \mid DocIndex_i \neq DocIndex_j\} \quad (16)$$

Accordingly, equation (16) is defined to create the negative sample by pairing captions of two images which have different document indices. For every two images $i$ and $j$ with different document indices in a batch, a negative sample can be created by pairing TrueCap$_i$ with TrueCap$_j$.

$$loss^{CapCapCoh2} = - \sum_{(TrueCap_i, TrueCap_j) \in CapCapPos} \log \sigma(MLP^{CapCapCoh2}([s_{L,</S>}^{True^{(i)}}, s_{L,</S>}^{True^{(j)}}]))$$
$$- \sum_{(TrueCap_i, TrueCap_j) \in CapCapNeg2} (1 - \log \sigma(MLP^{CapCapCoh2}([s_{L,</S>}^{True^{(i)}}, s_{L,</S>}^{True^{(j)}}]))) \quad (17)$$

Then, equation (17) is defined to compute the corresponding contrastive loss **$loss^{CapCapCoh2}$**. In the equation, the decoding state $s_{L,</S>}^{True^{(i)}}$ and $s_{L,</S>}^{True^{(j)}}$ for two captions of the positive samples are concatenated to compute the logits for the positive samples. And the decoding state $s_{L,</S>}^{True^{(i)}}$, $s_{L,</S>}^{True^{(j)}}$ for two captions of the corresponding negative samples are concatenated to compute the logits for the negative samples.

### 4.5 Training

The above sections have defined the generative loss, the contrastive loss of the caption-image-text coherence mechanism, and two types of contrastive losses of the caption-caption coherence mechanisms. The final loss can be defined by linearly combining the losses together as in equation (18).

$$loss = \lambda^{Gen} loss^{Gen} + \lambda^{CapITCoh} loss^{CapITCoh} + \lambda^{CapCapCoh1} loss^{CapCapCoh1} + \lambda^{CapCapCoh2} loss^{CapCapCoh2} \quad (18)$$

In equation (18), $\lambda^{Gen}$, $\lambda^{CapITCoh}$, $\lambda^{CapCapCoh1}$, and $\lambda^{CapCapCoh2}$ are four hyper-parameters. Empirically, $\lambda^{Gen}$, $\lambda^{CapITCoh}$, $\lambda^{CapCapCoh1}$, and $\lambda^{CapCapCoh2}$ are set as 1, 0.01, 0.01 and 0.1 respectively to consider all the coherence mechanisms. Setting $\lambda$ as 0 means not considering the corresponding coherence mechanism.

To make there are enough images from same documents and different documents for training caption-caption coherence mechanisms, the following algorithm is employed to create a batch. At each time, a document is randomly selected, and images in a document are ordered as their occurring orders. Then at most $W$ images with continuous indices are selected and added to the batch, and are removed from the document. The above steps continue until the number of images in the batch reaches BatchSize.

### 4.6 The two-level beam search algorithm

The two-level beam search algorithm is proposed to generate coherent captions based on the proposed model. Both generative loss and contrastive loss are used to score the generated captions. The generative loss is computed each time a word is generated, and the contrastive loss is computed when a complete caption is generated. The batch contains images within the window $W$ in a document, and the two-level beam search algorithm is applied to generate coherent captions for the $W$ images in the batch.

Fig. 3 shows the details of the algorithm. The first step of the algorithm is to use the word-level beam search algorithm to generate $C$ candidate captions for each image in the batch. Each candidate caption gets a generative score and a CapITCoh score, and the two scores are then linearly combined as the single-caption score. The second step is to compute the CapCapCoh score for each possible sequence of candidate captions, which is computed by averaging CapCapCoh scores of every two neighboring images. There are $C^W$ candidate caption combinations, the final scores of which is computed by adding the average single-caption score and the average CapCapCoh score. Since the number of possible caption sequences is very large, the caption-level beam search algorithm is use to get the caption sequence with the highest final score.

---

Algorithm: Two-level beam search algorithm for generating coherent captions

Inputs: $W$ images in a same document, contexts are denoted as $\{X_i^{Txt}, X_i^{Img}, X_i^{Face}, X_i^{Obj} \mid 1 \leq i \leq W\}$, BeamSize

Outputs: Coherent captions for the $W$ images

Steps:

    #Step 1: Word-level beam search algorithm for generating single captions

    For each image $i$:

        # Using the word-level beam search algorithm to generate $C$ captions with highest scores.

        # $Cap_{i,j}$ is the j-th generated caption for the i-th image, and $GenScore_{i,j}$ is the corresponding score.

        $\{(Cap_{i,1}, GenScore_{i,1}), \ldots, (Cap_{i,C}, GenScore_{i,C})\}$ = WordLevelBeamSearch($X_i^{Txt}, X_i^{Img}, X_i^{Face}, X_i^{Obj}$)

        For $j$ in range ( $C$ ):

            # the special token </S> is appended to the generated caption to compute the CapITCoh score

            $CapITCohScore_{i,j}$ = CapITCohMechanism ( $[Cap_{i,j}, </S>], X_i^{Txt}, X_i^{Img}, X_i^{Face}, X_i^{Obj}$ )

            $SingleCaptionScore_{i,j}$ = $GenScore_{i,1}$ + $CapITCohScore_{i,j}$

    # Step 2: Caption-level beam search algorithm for generate coherent caption sequences

    Beam = [{ }] # Save caption sequences

    For $i$ in Range ( $W$ ):

        CandiateBeam = []

        For each caption sequence CapSeq in Beam:

            For j in Range ( $C$ ):

                CapSeq = CapSeq $\cup$ [ $Cap_{i,j}$ ]

                # Averaging CapCapCoh scores of every two adjacent images

                AvgCapCapCohScore = ComputeAvgCapCapCohScore ( CapSeq )

                AvgSingleCaptionScore = averaging the SingleCaptionScore of all captions in CapSeq

                CapSeqScore = AvgSingleCaptionScore + AvgCapCapCohScore

                CandiateBeam = CandiateBeam $\cup$ [{CapSeq, CapSeqScore }]

        Beam = BeamSize caption sequences in CandiateBeam with top values of CapSeqScore

    Return the caption sequence with the highest CapSeqScore in Beam

**Fig. 3 The two-level beam search algorithm for generating coherent captions**

## 4.7 Definition of coherence evaluation metrics

Two types of caption-caption coherence mechanisms are proposed based on the coherence mechanism as follows. The first caption-caption coherence metrics is named **CapCapCohScore1** which is trained with the loss defined in equation (18) by setting $\lambda^{Gen}=0$, $\lambda^{CapITCoh}=0$, $\lambda^{CapCapCoh1}=1$, and $\lambda^{CapCapCoh2}=0$, corresponding to the first caption-caption coherence mechanism CapCapCoh1. The second caption-caption coherence metrics is named **CapCapCohScore2** which is trained with the loss defined in equation (18) by setting $\lambda^{Gen}=0$, $\lambda^{CapITCoh}=0$, $\lambda^{CapCapCoh1}=0$, and $\lambda^{CapCapCoh2}=1$, corresponding to the second caption-caption coherence mechanism CapCapCoh2.

$$CapCapCohScore1(cap_1, cap_2) = \sigma(MLP^{CapCapCoh1}([s_{L,</S>}^{cap1}, s_{L,</S>}^{cap2}])) \times 100 \quad (19)$$

$$CapCapCohScore2(cap_1, cap_2) = \sigma(MLP^{CapCapCoh2}([s_{L,</S>}^{cap1}, s_{L,</S>}^{cap2}])) \times 100 \quad (20)$$

For two captions $cap_1$ and $cap_2$ of two neighboring images, **CapCapCohScore1** and **CapCapCohScore2** can be computed using equations (19) and (20) respectively. Given a sequence of images, the scores of CapCapCohScore1 and CapCapCohScore2 of the sequence are computed by averaging the corresponding scores of all pairs of 1-hops neighboring images.

## 5 Experiment setups

### 5.1 Implementation and parameter settings

The codes of training and inference pipelines are written with PyTorch using the AllenNLP framework [37]. The pre-trained encoder RoBERTa is adapted from fairseq [38]. Training is carried out with mixed precision to reduce the memory footprint and allow the proposed full model to be trained on a single GPU. The full model takes 5 days to train on an RTX 3090 GPU and has more than 200 million trainable parameters. The codes and the datasets will be available in Github via the link https://github.com/jingqiangchen/ConCaps.

The parameter settings of the coherence mechanisms are set as follows. The window size $W$ for selecting neighboring images is set as 3. The batch size is set as 15, and the beam size for the two-level beam search algorithm is set as 3. The hidden sizes of the MLPs of the coherence mechanisms are set as 1024. The parameter settings of the generation model are set following [4]. The number of heads is set as 16. The hidden size is set as 1024. For parameter optimization, the adaptive gradient algorithm Adam [39] with the parameter settings $\beta_1$=0.9, $\beta_1$=0.9, $\varepsilon$=10$^{-6}$ is used. The learning rate is warmed up in the first 5% of the training steps to 10$^{-4}$, which is decayed linearly in the training.

**5.2 Methods for comparisons**

Four variations of the proposed method are compared with the following baseline methods.

**TextRank** [40] is a state-of-the-art graph-based extractive summarization method, which only uses the accompanying text of the image as input to extract sentences as the caption.

**Show Attend Tell** [2] is the state-of-the-art neural image captioning method that uses the attentional encoder-decoder model by attending to image splits for caption generation. This baseline method only uses the image as input.

**Pooled Embeddings and Tough-to-beat** [5] are template-based models that encode documents at the sentence level and pay attention to certain sentences at different time steps. Pooled Embeddings computes sentence representations by averaging word embeddings and adopts context insertion in the second state. The tough-to-beat method obtains sentence representations from the tough-to-beat method [41] and uses sentence-level attention weights to insert named entities.

**Transform Tell** [4] is the transformer-based attention model using a pre-trained RoBERTa as the text encoder and a transformer as the decoder. This baseline method is the generation model of the proposed method in this paper.

**Visual News Captioner** is proposed in [6], which is based on Transformer. This baseline method adopts Multi-Head Attention on Attention. Named entities are added as another text source to help predict named entities more accurately. The visual selective layer is proposed to strengthen the connection between the image and text. The multi-head pointer-generator module is adopted as the generation model. The tag-cleaning operation is proposed to handle the OOV problem.

**MMKG Captioner** [7] constructs a multi-modal knowledge graph to associate the visual objects with named entities and capture the relationship between entities simultaneously with the help of external knowledge. A text sub-graph is built by extracting named entities and their relationships from the article, and an image sub-graph is built by detecting the objects in the image. These two sub-graphs are connected with a cross-modal entity matching module. Finally, the multi-modal knowledge graph is integrated into the captioning model to generated captions.

**CohCaps** is the proposed method of this paper, which incorporates coherence mechanisms into the single-image captioning model to generate coherent captions for multiple images. This model consists of a generation model and two types of coherence mechanisms. The caption-image-text coherence mechanism models the coherence between the caption, the image and the accompanying text, and the caption-caption coherence mechanisms models the coherence among neighboring images in a document. Four variations of CohCaps with different coherence mechanisms are compared. CohCaps+CapITCoh+CapCapCoh1&2 is with all the coherence mechanisms. CohCaps+CapITCoh+CapCapCoh1 is without CapCapCoh2 by setting $\lambda^{CapCapCoh2}=0$. CohCaps+CapITCoh+CapCapCoh2 is without CapCapCoh2 by setting $\lambda^{CapCapCoh1}=0$. CohCaps+CapITCoh is without CapCapCoh1 and CapCapCoh2 by setting $\lambda^{CapCapCoh1}=0$ and $\lambda^{CapCapCoh2}=0$.

## 6 Evaluations of single-image captioning

To see the performance of the proposed model on single-image captioning, and whether the proposed coherence mechanisms can improve entity-aware image captioning, six metrics i.e. BLEU-4 [41], METEOR [42], ROUGE[30], CIDEr [43], Named Entity Precision (NE Precision), Named Entity Recall (NE Recall) are adopted. Among these metrics, NE Precision and NE Recall are the Precision and Recall scores of named entities in the generated captions with regard to the ground-of-the-truth captions. The implementation of the metrics is obtained using the COCO evaluation toolkit. The precision and recall on named entities are computed by comparing named entities contained in the generated captions and the ground-of-the-truth captions as the evaluation metrics. Named entities are identified in the generated captions and the ground-of-the-truth captions using SpaCy.

Table 3, Table 4, and Table 5 show the evaluation results on DM800K, NYTimes800K, and GoodNews respectively. The scores of the 6 baseline methods on NYTimes800K and GoodNews have been reported in [4] and [5]. For the new dataset DM800K, three baselines are run on the dataset and the scores are reported. The scores of four variations of the proposed method CohCaps with different combinations of coherence mechanisms are reported.

According to Table 3, CohCaps outperforms TextRank, Show Attend Tell, and Transform Tell for all the 6 metrics on DM800K. According to Table 4, CohCaps outperforms the 7 baseline methods for all the 6 metrics on NYTimes800K. According to Table 5, CohCaps outperforms all the baselines for the three metrics BLUE-4, METEOR and ROUGE on GoodNews. In particular, CohCaps outperforms the baseline method Transform Tell on all the three datasets for all the 6 evaluation metrics. Note that Transform Tell is similar to the generation model of the proposed CohCaps. The proposed coherence mechanisms are the main difference between Transform Tell and CohCaps. According to the scores, the coherence mechanisms can improve captioning models by adding the coherence-related features to the captioning model.

Table 3  Evaluations of single-image captioning on DM800K

| Method | BLEU-4 | METEOR | ROUGE | CIDEr | NE Precision | NE Recall |
|---|---|---|---|---|---|---|
| TextRank | 1.3 | 12.9 | 13.4 | 1.4 | 9.6 | 32.9 |
| Show Attend Tell | 0.5 | 4.4 | 15.4 | 4.5 | - | - |
| Transform Tell | 6.5 | 10.1 | 17.6 | 43.1 | 18.7 | 17.6 |
| CohCaps | | | | | | |
| +CapITCoh | 6.6 | 10.1 | 17.7 | 43.1 | 18.7 | 17.7 |
| +CapITCoh+CapCapCoh1 | **6.8** | **10.2** | 17.8 | 43.3 | **18.8** | **17.8** |
| +CapITCoh+CapCapCoh2 | 6.7 | 10.1 | **17.8** | 43.4 | 18.7 | 17.8 |
| +CapITCoh+CapCapCoh1&2 | 6.5 | **10.1** | 17.6 | **43.5** | 18.5 | 17.6 |

Bold values indicate that the best results

Table 4  Evaluations of single-image captioning on NYTimes800K

| Method | BLEU-4 | METEOR | ROUGE | CIDEr | NE Precision | NE Recall |
|---|---|---|---|---|---|---|
| TextRank | 1.9 | 7.3 | 11.4 | 9.8 | 3.6 | 4.9 |
| Show Attend Tell | 0.9 | 4.3 | 13.9 | 4.7 | - | - |
| Tough-to-beat | 0.7 | 4.2 | 11.5 | 12.5 | 8.9 | 7.7 |
| Pooled Embeddings | 0.8 | 4.1 | 11.3 | 12.2 | 8.6 | 7.3 |
| Visual News Captioner | 6.4 | 8.1 | 21.9 | 56.1 | 24.8 | 22.3 |
| Transform Tell | 6.3 | 10.4* | 21.7 | 54.4 | 24.6 | 22.2 |
| MMKG Captioner | 6.3 | 6.6 | 21.6 | 54.0 | - | - |
| CohCaps | | | | | | |
| +CapITCoh | 6.4 | 10.4 | 21.7 | 54.7 | 24.3 | 22.4 |
| +CapITCoh+CapCapCoh1 | 6.4 | 10.5 | 21.7 | 54.5 | 24.4 | 22.4 |
| +CapITCoh+CapCapCoh2 | 6.4 | 10.5 | **21.9** | 55.0 | 24.5 | 22.5 |
| +CapITCoh+CapCapCoh1&2 | **6.6** | **10.6** | 21.8 | 55.2 | 24.6 | **22.5** |

Bold values indicate that the best results. The score with * is not reported in the original paper, and we compute the score.

Table 5  Evaluations of single-image captioning on GoodNews

| Method | BLEU-4 | METEOR | ROUGE | CIDEr | NE Precision | NE Recall |
|---|---|---|---|---|---|---|
| TextRank | 1.7 | 7.5 | 11.6 | 9.5 | 1.7 | 5.1 |
| Show Attend Tell | 0.7 | 4.1 | 11.9 | 12.2 | - | - |
| Tough-to-beat | 0.8 | 4.2 | 11.8 | 12.8 | 9.1 | 7.8 |
| Pooled Embeddings | 0.8 | 4.3 | 12.1 | 12.7 | 8.2 | 7.2 |
| Visual News Captioner | 6.1 | 8.3 | 21.6 | 55.4 | 22.9 | 19.3 |
| Transform Tell | 6.0 | 10.2* | 21.4 | 53.8 | 22.2 | 18.7 |
| MMKG Captioner | 6.1 | 6.3 | 21.5 | 54.0 | - | - |
| CohCaps | | | | | | |
| +CapITCoh | 6.1 | 10.2 | 21.4 | 54.2 | 22.2 | 18.7 |
| +CapITCoh+CapCapCoh1 | 6.1 | 10.2 | 21.5 | 54.2 | 22.4 | 18.8 |
| +CapITCoh+CapCapCoh2 | 6.1 | 10.2 | 21.5 | 54.3 | 22.3 | 18.8 |
| +CapITCoh+CapCapCoh1&2 | **6.2** | **10.3** | **21.7** | 54.3 | 22.2 | 18.7 |

Bold values indicate that the best results. The score with * is not reported in the original paper, and we compute the score.

CohCaps+CapITCoh performs as well as Transform Tell, indicating that adding the caption-image-text coherence mechanism to the captioning model will at least not decrease the performance of the model. The reason is that the captions generated by the captioning model without the caption-image-text coherence mechanism are inherently coherent with the images and the accompanying texts, and there are limited room for the coherence mechanism to improve the captioning model.

The models CohCaps with caption-caption coherence mechanisms perform better than the model CohCaps without caption-caption coherence mechanisms. The reason is as follows. The captioning model without caption-caption coherence mechanisms generates captions for each single image independently, not considering the relationships between images. In contrast, the models CohCaps with caption-caption coherence mechanisms bring the features of the coherence relations between neighboring images into the captioning model. These features are orthogonal to the captioning model and can improve the model. Moreover, the two caption-caption coherence mechanisms are two different features for the captioning model, and can improve the captioning model differently. CohCaps+CapITCoh+CapCapCoh1 performs better on DM800K, and CohCaps+CapITCoh1+CapCapCoh2 perform better on NYTimes800K and GoodNews.

CohCaps performs better in DM800K than in the other two datasets. The reason is that DM800K contains more images per document. As shown in Table 1, there are average 4.45 images per document in DM800K, about three times that of NYTimes800K and GoodNews. More images in a document means there is more coherence information of images to be mined to improve the captioning model.

## 7 Coherence evaluations of multi-image captions

### 7.1 Automatic evaluations

To see performance of the proposed captioning model on multi-image captioning, the two proposed coherence metrics CapCapCohScore1 and CapCapCohScore2 defined in Section 4.7 are adopted to evaluate coherence between generated captions of multiple images. The two metrics Caption Rouge Score and Caption Entity Score defined in Section 3.2 are also computed for the generated captions. Table 6, Table 7 and Table 8 shows the scores for DM800K, NYTimes800K and GoodNews.

In three tables, the upper bound is obtained by computing the coherence scores on the ground-of-the-truth captions, and the lower bounds are obtained by computing the coherence scores on the baseline methods Show Attend Tell and Transform Tell. Four variations of CohCaps are evaluated.

**Table 6** Coherence evaluations on DM800K

| Method | CapCapCohScore1 | CapCapCohScore2 | Caption Rouge Score | Caption Entity Score |
|---|---|---|---|---|
| Gound of Truth (Upper Bound) | **13.83** | 9.98 | 20.43 | 0.2199 |
| Show Attend Tell | 12.86 | 9.09 | 27.62 | - |
| Transform Tell | 13.32 | 9.27 | 44.28 | 0.4084 |
| CohCaps | | | | |
|   +CapITCoh | 13.33 | 9.25 | 45.88 | 0.4197 |
|   +CapITCoh+CapCapCoh1 | 13.45 | 9.43 | 46.01 | 0.4204 |
|   +CapITCoh+CapCapCoh2 | 13.43 | **9.48** | 45.95 | 0.4217 |
|   +CapITCoh+CapCapCoh1&2 | **13.47** | 9.43 | 45.52 | 0.4172 |

Bold values indicate the best results.

**Table 7** Coherence evaluations on NYTimes800K

| Method | CapCapCohScore1 | CapCapCohScore2 | Caption Rouge Score | Caption Entity Score |
|---|---|---|---|---|
| Gound of Truth (Upper Bound) | **17.09** | **8.49** | 12.31 | 0.1197 |
| Show Attend Tell | 15.38 | 7.85 | 22.10 | - |
| Transform Tell | 16.04 | 8.03 | 27.41 | 0.2736 |
| CohCaps | | | | |
|   +CapITCoh | 16.06 | 8.02 | 27.69 | 0.2746 |
|   +CapITCoh+CapCapCoh1 | **16.19** | 8.15 | 27.59 | 0.2757 |
|   +CapITCoh+CapCapCoh2 | 16.15 | **8.17** | 28.02 | 0.2781 |
|   +CapITCoh+CapCapCoh1&2 | 16.15 | 8.14 | 27.81 | 0.2771 |

Bold values indicate the best results.

**Table 8** Coherence evaluations on GoodNews

| Method | CapCapCohScore1 | CapCapCohScore2 | Caption Rouge Score | Caption Entity Score |
|---|---|---|---|---|
| Gound of Truth (Upper Bound) | **25.05** | **13.32** | 13.56 | 0.1342 |
| Show Attend Tell | 22.42 | 10.70 | 28.04 | - |
| Transform Tell | 24.47 | 12.72 | 30.68 | 0.3070 |
| CohCaps | | | | |
|   +CapITCoh | 24.85 | 12.56 | 32.70 | 0.3395 |
|   +CapITCoh+CapCapCoh1 | 25.44 | **12.84** | 33.44 | 0.3458 |
|   +CapITCoh+CapCapCoh2 | 25.33 | 12.72 | 32.54 | 0.3282 |
|   +CapITCoh+CapCapCoh1&2 | **25.48** | 12.76 | **33.55** | **0.3462** |

Bold values indicate the best results.

The most surprising thing in Table 6, Table 7 and Table 8 is that the Caption Rouge Scores and the Caption Entity Scores of the Ground of The Truth captions are much lower than that of the automatically generated captions in both DM800K and NYTimes800K datasets. The scores of the Ground of The Truth are in line with the scores of the training set shown in Table 2, which indicates that the two scores are applicable for evaluations among Ground of The Truth captions. However, the scores for the automatically generated captions tend to be higher. This is because all the generation models in the tables use the accompanying texts of the images to generate captions, and neighboring images have similar accompanying texts, leading the generation models to generate captions with overlapped text from the accompanying texts. In contrast, the Ground of The Truth captions are manually written captions which have less text overlaps than automatically generated captions. Caption Rouge Score and Caption Entity Score are computed to reflect surface text overlaps of captions, making the scores of the Ground of The Truth lower than that of the automatically generated captions. The CapCapCohScore1 and CapCapCohScore1 scores for GoodNews are much higher than that of the other two datasets. This is mainly because the leading 512 words in the document are used as accompanying texts for all the images in the document in GoodNews to compute the scores. As mentioned in Section 3.2, image positions are not provided in GoodNews.

The proposed metrics CapCapCohScore1 and CapCapCohScore2 perform better than Caption Rouge Score and the Caption Entity Score when evaluating coherence of the Ground of Truth and the generated captions. In Table 6, Table 7 and Table 8, CapCapCohScore1 and CapCapCohScore2 of the Ground of Truth are much higher than that of the automatically generated captions. The two scores are computed by counting semantic relationships of images, captions and accompanying texts for two neighboring images through the proposed coherence mechanisms. This is different from Caption Rouge Score and the Caption Entity Score which only compute text overlaps of two captions.

According to the three tables, CohCaps outperforms the baselines Show Attend Tell and Transform Tell. Note that the generation model of CohCaps is borrowed from Transform Tell. Moreover, the four variations of CohCaps perform differently. The scores of CohCaps+CapITCoh are not higher than that of the baseline method Transform Tell. CohCaps with the caption-caption coherence mechanisms outperforms the model without caption-caption coherence mechanisms. This indicates that caption-caption coherence mechanisms can improve coherence of generated captions by considering coherence relations in the model.

The two caption-caption coherence metrics perform different in the two datasets. The CapCapCohScore1 scores in NYTimes800K are higher than that of DM800K, while the CapCapCohScore2 in NYTimes800K are lower than that of DM800K. The two datasets are different datasets, and the coherence mechanisms are trained on the datasets independently. For either dataset, the ranking of the CapCapCohScore1 scores are in line with the ranking of the CapCapCohScore2 scores.

**7.2 Human evaluations**

Qualitative human evaluations are carried out to show whether caption-caption coherence mechanisms improve multi-image captioning, and whether human evaluations are in line with automatic evaluations. 30 documents are randomly sampled from DM800K, each of which contains first 3 images. Three volunteers are asked to evaluate the captions generated by three methods: Transform Tell, CohCaps+CapITCoh+CapCapCoh1, CohCaps+CapITCoh+CapCapCoh2. The names of the methods are hidden in the evaluation for the sake of fair comparisons. The volunteers are asked to rank the generated captions from the three aspects: informativeness, coherence, overall quality. The best will be given the value 1, the worse will be given the value 2, and the worst will be given the value 3.

Table 9  Human evaluation results

| Method | Informativeness | Coherence | Overall |
| --- | --- | --- | --- |
| Transform Tell | 2.08 | 2.05 | 2.08 |
| CohCaps | | | |
| +CapITCoh+CapCapCoh1 | 1.98 | **1.92** | **1.94** |
| +CapITCoh+CapCapCoh2 | **1.95** | 2.02 | 1.98 |

Bold values indicate that the best results

Table 9 shows the results of human evaluation. CohCaps with the caption-caption coherence mechanisms are better than Transform Tell in all three aspects, especially in coherence. The result is in line with the evaluation results of the metrics CapCapCohScore1 and CapCapCohScore2. This result indicates that coherence mechanisms can improve coherence of the generated captions, as well as the overall quality.

## 8  Case study of the coherence metrics

The examples in the Introduction section are used to demonstrate what the proposed coherence metrics represents.

Table 10 shows the CapCapCohScore1 and the CapCapCohScore2 of the ground-of-the-truth captions of the images in Fig. 1, the captions generated by Transform Tell, and the captions generated by three variations of the proposed method CohCaps. To show how the scores are computed, we take the CapCapCoh Score1 26.23 of the three ground-of-the-truth captions for example. The Cap2CapCoh score of the first caption and the second caption, and the CapCapCoh score of the second caption and the third caption are computed respectively. And 26.23 is the average of the two scores. To compute the CapCapCoh score of the first caption cap1 and the second caption cap2, the two corresponding images are encoded to get image encodings, object encodings, and face encodings, and the accompanying text are encoded to get text encodings. Then equations (7) to (9) are applied to obtain $s_{L,</S>}^{cap1}$ and $s_{L,</S>}^{cap2}$. Finally, equations (19) and (20) are applied to compute CapCapCohScore1 and CapCapCohScore2.

The ground-of-the-truth captions obtain the highest CapCapCohScore2 score. While it is surprising that the CapCapCohScore1 scores of CohCaps+CapITCoh+CapCapCoh1 and CohCaps+CapITCoh+CapCapCoh2 are higher than that of the ground-of-the-truth caption. Note that CapCapCoh2 is trained by replacing entities or nouns of the ground-of-the-truth captions with same-typed randomly selected entities or nouns, which are contrasted with the ground-of-the-truth captions of neighboring images. This can enhance coherence of captions with same entities or nouns. The ground-of-the-truth captions of the three images have the same noun spacecraft, while the captions generated by CohCaps+CapITCoh+CapCapCoh1 and CohCaps+CapITCoh+CapCapCoh2 have two same entities and nouns i.e. spacecraft and Jupiter. In contrast, only two of the captions generated by the baseline method Transform Tell have the same noun spacecraft.

**Table 10** Generated captions and scores of three images of the example in Fig. 1

| Method | Captions | CapCapCoh Score1 | CapCapCoh Score2 |
|---|---|---|---|
| Gound of Truth | The **spacecraft** will take 5 five years to reach orbit around **Jupiter** and then spend about one year gathering information on **Jupiter**' magnetic field, atmosphere and interior | 26.23 | 15.39 |
| | The windmill-shaped **spacecraft** is equipped with three tractor-trailer-size solar panels for its two billion-mile journey into the outer solar system | | |
| | Each of Juno's three wings is 29ft long and 9ft wide. The panels - folded for launch - are here seen emanating from the **spacecraft** much like the blades of a windmill | | |
| Transform Tell | The Juno **spacecraft**, pictured, is seen in this artist's impression of the launch pad at Cape Canaveral Air Force Station | 20.11 | 14.32 |
| | The mission will be launched in the next five years, with the first ever mission to Mars orbit and back for a fly-by of the Earth in 2013 | | |
| | The Juno **spacecraft** will be able to take a trip to Jupiter in 2013, which is nearly 20,000 times as powerful as Earth's | | |
| CohCaps +CapITCoh +CapCapCoh1 | The Juno **spacecraft**, seen here in a photo from the launch pad, was launched on a path to **Jupiter** | 34.29 | 14.78 |
| | The **spacecraft** will be launched in the next five years to **Jupiter**, and will be powered only by solar energy, making it the farst-travelled **spacecraft** ever driven by the Sun | | |
| | The Juno **spacecraft** will get closer to **Jupiter** than any previous mission, making its closest approach 3,100 miles above the planet's cloud tops | | |
| CohCaps +CapITCoh +CapCapCoh2 | The Juno **spacecraft**, seen here in this artist's impression, was launched on a path to **Jupiter** | 28.40 | 14.66 |
| | The six-ton rocket will be used to take it to the Earth's gravity and slingshot itself further out towards **Jupiter** | | |
| | The success **spacecraft** will be launched from **Jupiter's** orbit around the planet | | |
| CohCaps +CapITCoh +CapCapCoh1&2 | The **spacecraft**, pictured, was launched from Cape Canaveral Air Force Station in Florida on Thursday | 20.35 | 14.54 |
| | The **spacecraft**, pictured, launched by Nasa in July, will be launched in the next five years | | |
| | It will be the fastest-travelling man-made object ever | | |

To see how are the scores of fake captions, some of the words spacecraft in the captions generated by CohCaps+CapITCoh+CapCapCoh1 is replaced by the noun machine, and three groups of fake captions are obtained. Table 11 shows the scores. For the first group of fake captions, only the word in the caption of the second image is replaced with the word machine. For the second group of fake captions, the word spacecraft in the captions of all the three images are replaced. And for the third group of fake captions, the word spacecraft in the captions of the first two images are replaces. The scores of the three captions groups are all lower than that of the original caption group. The second group of fake captions gain the highest scores among the three fake caption group, mainly because all the word spacecraft in the captions replaced with the word machine, making the captions be coherent with each other. In contrast, parts of the word spacecraft of the first group and the third group are replaced.

**Table 11** Scores of fake captions for three images of the example in Fig. 1

| Method | Captions | CapCapCoh Score1 | CapCapCoh Score2 |
|---|---|---|---|
| Fake captions 1 | The **spacecraft**, seen here in a photo from the launch pad, was launched on a path to **Jupiter**<br>The **machine** will be launched to **Jupiter** in the next five years, and will be powered only by solar energy, making it the farst-travelled machine ever driven by the Sun<br>The Juno **spacecraft** will get closer to **Jupiter** than any previous mission, making its closest approach 3,100 miles above the planet's cloud tops | 29.75 | 14.32 |
| Fake captions 2 | The **machine**, seen here in a photo from the launch pad, was launched on a path to **Jupiter**<br>The **machine** will be launched to **Jupiter** in the next five years, and will be powered only by solar energy, making it the farst-travelled machine ever driven by the Sun<br>The Juno **machine** will get closer to **Jupiter** than any previous mission, making its closest approach 3,100 miles above the planet's cloud tops | 34.01 | 14.40 |
| Fake captions 3 | The **machine**, seen here in a photo from the launch pad, was launched on a path to Jupiter<br>The **machine** will be launched to **Jupiter** in the next five years, and will be powered only by solar energy, making it the farst-travelled machine ever driven by the Sun<br>The Juno **spacecraft** will get closer to **Jupiter** than any previous mission, making its closest approach 3,100 miles above the planet's cloud tops | 29.96 | 14.30 |

## 9 Conclusions and Future Work

This paper explores the task of entity-aware multi-image captioning. This task is different from entity-aware single-image captioning in that the former has to consider coherence relationships between the caption, the image and the accompanying text, i.e. caption-image-text coherence, but also has to consider coherence relationships between neighboring images in a document, i.e. caption-caption coherence. Caption-image-text coherence renders entities in the accompanying text of the image be also in the caption. Caption-caption coherence renders entities in the caption of the image be also in the caption of the caption of its neighboring image. Moreover, existing datasets for single-image captioning contain relatively less images per document. To tackle the task, this paper constructs the new dataset DM800K that contains more images per document, and proposes the coherent entity-aware multi-image captioning model (CohCaps) by utilizing the two coherence relationships. CohCaps consists of the Transformer-based caption generation model, the contrastive caption-image-text coherence mechanism (CapITCoh), and the contrastive caption-caption coherence mechanism (CapCapCoh). The generation model encodes the accompanying text and the image, and then generates captions by paying attention to the encodings. The contrastive caption-image-text coherence mechanism aims to make the generated caption be coherent with the image and the accompanying, and is trained by contrasting true captions with fake captions. The contrastive caption-caption coherence mechanism aims to make the generated caption be coherent with captions of neighboring images, and is trained by contrasting positive image caption pairs and negative image caption pairs. Two coherence metrics are proposed based on the caption-caption coherence mechanisms to evaluate coherence between captions of neighboring images. Experiments on three datasets show the proposed method CohCaps outperforms 7 baseline methods, indicating considering coherence relationships in the captioning model can improve single-image captioning. Experiments also show that the generated captions of CohCaps are more coherent than that of baselines according to automatic evaluations and human evaluations.

As our future work, we will exploit other relationships for entity-aware multi-image captioning, such as image-image similarity, accompanying text-text similarity, and etc.

**Declaration of competing interest**

The authors declare that they have no known competing financial interests or personal relationships that could have appeared to influence the work reported in this paper.

**Data availability**

Data will be made available on request

**Acknowledgements**